\documentclass{article}

\usepackage{microtype}
\usepackage{graphicx}
\usepackage{subfigure}
\usepackage{balance}
\usepackage{booktabs} 
\usepackage{hyperref}

\usepackage[arxiv]{icml2020}
\usepackage{amsmath}
\icmltitlerunning{Salvaging Federated Learning by Local Adaptation}

\newcommand{\paragraphbe}[1]{\vspace{0.75ex}\noindent{\bf \em #1} }

\begin{document}
\twocolumn[
\icmltitle{Salvaging Federated Learning by Local Adaptation}

\begin{icmlauthorlist}
\vspace{-1em}
\icmlauthor{Tao Yu}{cornell}
\icmlauthor{Eugene Bagdasaryan}{cornell}
\icmlauthor{Vitaly Shmatikov}{cornell}
\vspace{1em}
\end{icmlauthorlist}
\icmlaffiliation{cornell}{Department of Computer Science, Cornell University.}

\icmlcorrespondingauthor{Tao Yu}{tyu@cs.cornell.edu}
\icmlcorrespondingauthor{Eugene Bagdasaryan}{eugene@cs.cornell.edu}
\icmlkeywords{Machine Learning, Federated Learning, Local Adaptation, Privacy, Robustness}
]

\printAffiliationsAndNotice{}
\begin{abstract}
\label{sec:abstract}

Federated learning (FL) is a heavily promoted approach for training ML
models on sensitive data, e.g., text typed by users on their smartphones.
FL is expressly designed for training on data that are unbalanced and
non-iid across the participants.  To ensure privacy and integrity of
the fedeated model, latest FL approaches use differential privacy or
robust aggregation.

We look at FL from the \emph{local} viewpoint of an individual participant
and ask: (1) do participants have an incentive to participate in FL? (2) how can participants \emph{individually} improve the quality
of their local models, without re-designing the FL framework and/or
involving other participants?

First, we show that on standard tasks such as next-word prediction,
many participants gain no benefit from FL because the federated model
is less accurate on their data than the models they can train locally
on their own.  Second, we show that differential privacy and robust
aggregation make this problem worse by further destroying the accuracy
of the federated model for many participants.

Then, we evaluate three techniques for local adaptation of federated
models: fine-tuning, multi-task learning, and knowledge distillation. We
analyze where each is applicable and demonstrate that all participants
benefit from local adaptation.  Participants whose local models are poor
obtain big accuracy improvements over conventional FL.  Participants whose
local models are better than the federated model\textemdash and who
have no incentive to participate in FL today\textemdash improve less,
but sufficiently to make the adapted federated model better than their
local models.

\end{abstract}

\section{Introduction}
\label{sec:introduction}

Federated learning~\cite{fedlearn_1} is a framework for large-scale,
distributed learning on sensitive data, e.g., training a next-word
prediction model on texts typed by users into their smartphones or a
medical treatment model on patient records from multiple hospitals.
Designed for unbalanced, non-iid data distributions, federated learning
has demonstrated good performance and scalability~\cite{fedlearn_sys}
and is promoted by Google~\cite{pichar-keynote} and other
companies as the solution to privacy problems in predictive
keyboards~\cite{hard2018federated}, medicine~\cite{docai}, and other
domains~\cite{kairouz2019advances}.

The original approach~\cite{fedlearn_1} creates the federated model by
repeatedly averaging model updates from small subsets of participants.
Both the updates and the final model can leak participants' training
data, violating privacy~\cite{shokri2017membership, melis2018inference}.
Averaging-based aggregation is also vulnerable to attacks on model
integrity because malicious participants can introduce unwanted
behavior into the model~\cite{bagdasaryan2018backdoor}.  To protect
privacy, differentially private federated learning~\cite{fedlearn_dp}
bounds how much the model can reveal about the inputs from
any individual participant.  To protect integrity, robust
aggregation~\cite{yin2018byzantine} replaces average with median to
bound the influence of outliers on the model.

Users have an incentive to participate in federated learning only
if federated models are more accurate than the models they can train
independently on their own data.  Privacy and robustness mechanisms
introduce a fundamental conflict into this reasoning.  To take advantage
of the data of the unusual participants\textemdash which is one of the
principal design objectives of federated learning\textemdash aggregation
must incorporate their contributions into the federated model.  To protect
privacy and integrity, aggregation must restrict these contributions
from having much influence on the federated model.


\paragraphbe{Our contributions.}
We look at federated learning (FL) from the local perspective of
individual participants and investigate whether they have an incentive
to participate.  Does federated learning yield more accurate models
for \emph{them}?  If no, what can they do \emph{locally} to improve the
quality of models they obtain from FL.

First, we demonstrate that privacy and robustness protections destroy
the accuracy of federated models for many participants, removing their
main incentive to join federated learning.  We use standard federated
learning tasks: next-word prediction and image classification.  With very
few exceptions (see Section~\ref{sec:relatedwork}), prior work focused
on measuring the overall accuracy of federated models.  By contrast,
we (a) measure their accuracy for the individual participants, and
(b) show that many participants gain no benefit because the federated
model achieves worse accuracy on \emph{their} data than a model they
can train independently.  For example, when training a word-prediction
model on a Reddit dataset, the federated model based on robust median
aggregation achieves worse accuracy than the local models for the majority
of participants.

Next, we solve this fundamental tradeoff between privacy/robustness and
individual accuracy.  Instead of a single model that should be accurate
for all participants, we use \emph{local adaptation} to convert the
federated model into individual models for each participant.

Crucially, we are interested in local methods that \emph{an individual
participant can deploy on their own}.  We do not aim to change FL
aggregation algorithms because FL frameworks are controlled by platform
operators such as Google and cannot be changed unilaterally by a single
participant (e.g., a single smartphone).  We are also not interested in
solutions that require all or most participants to change their algorithms
because they require cooperation and are hard to deploy in practice.

We investigate three adaptation mechanisms: fine-tuning, multi-task
learning, and knowledge distillation.  We analyze where each is
applicable, and show how local adaptation helps participants recover
the accuracy destroyed by differential privacy and robust aggregation.
Participants who had no incentive to join federated learning because
their local models are better than the federated model benefit because
the adapted federated model becomes better than the local models.
For example, for 80\% of the participants in the word-prediction task,
the adapted robust model outperforms their local models.  Participants
whose local models are inaccurate\textemdash and thus already benefit from
federated learning\textemdash experience the biggest accuracy improvements
due to local adaptation and benefit even further.  Finally, we relate
the effects of adaptation to the complexity of participants' data.

\section{Related Work}
\label{sec:relatedwork}


\noindent
\textbf{\em Privacy and integrity of federated learning.}
Participants' model updates leak their training
data~\cite{melis2018inference}, and malicious participants can inject
unwanted behaviors into the model~\cite{bagdasaryan2018backdoor,
bhagoji2018analyzing}.  Secure aggregation~\cite{bonawitz2017practical}
prevents the global server from observing individual updates, but it also
makes attacks on integrity impossible to detect and the final model may
still leak training data.

Federated learning with differential privacy~\cite{fedlearn_dp} limits
the leakage of training data.  To limit the influence of individual
participants, several robust, ``Byzantine-tolerant'' aggregation schemes
have been proposed~\cite{blanchard2017machine, mhamdi2018hidden,
damaskinos2019aggregathor, rajput2019detox, chen2017distributed}.
Alternative aggregation schemes~\cite{yurochkin2019bayesian, guha2019one,
hsu2019measuring} for various flavors of federated learning provide
neither privacy, nor robustness.  Peer-to-peer (not federated) learning
with convex losses and without robustness is studied in~\cite{bellet18}.


\paragraphbe{Accuracy for individual participants.}
Federated learning is explicitly designed for non-iid participants, but
most prior work does not measure their individual accuracy.  Training of
participant-specific models is studied in~\cite{smith17}, without privacy
or robustness and at the cost of replacing the entire federated learning
framework.  Differential privacy disproportionately reduces model accuracy
for underrepresented participants~\cite{bagdasaryan2019differential}.
No previous work investigated the impact of robust aggregation on
individual accuracy.

Prior work on personalization of ML models focused on speaker
adaptation of acoustic models (see~\cite{yu2017recent}).  Many
techniques are not compatible with federated learning because
they require an ensemble of models~\cite{tan2015cluster} or are
speech-specific~\cite{miao2015speaker}, but~\cite{huang2015rapid}
connects personalization and multi-task learning.


Recent papers on personalizing federated models~\cite{wang2019federated,
jiang2019improving, fallah2020personalized, dinh2020personalized}
propose various methods to improve the accuracy for individual
participants; \cite{jiang2019improving} also connects meta learning
with personalization.  These papers do not investigate (a) if federated
models are more accurate than the models individual participants can
train on their own, (b) the impact of privacy and integrity protections
on individual participants' accuracy, and (c) purely local adaptation
techniques other than fine-tuning (the only exception is a short
paper~\cite{peterson2019private} that uses domain adaptation to counteract
the reduction in accuracy due to differential privacy).  While there
are dozens of alternative aggregation algorithms that may improve the
quality of local models (see the survey in~\cite{kairouz2019advances}),
all of them involve global changes to the federated learning framework,
require all participants to replace their algorithms, and cannot be
deployed locally and unilaterally by a participant.  As explained
in Section~\ref{sec:introduction}, \textbf{we are interested in
\underline{local} techniques that an individual participant can use to
mitigate the damage from privacy and robustness mechanisms}.

\section{Background}
\label{sec:background}




\noindent
\textbf{\em Federated learning} is a distributed learning paradigm for
training a model on multiple participants' data~\cite{fedlearn_1}.
The global server creates the initial model $G^0$.  In each round
$t=1..T$, the server selects a subset of $m$ participants from some pool
$Q$ of size $n$ and sends them the current model $G^{t-1}$.  Each selected
participant $i \in m$ updates the model on his local data $\mathcal{D}_i$
and sends the resulting model $P^{t}_i$
to the global server, which averages it with the other updates using the
aggregation learning rate $\eta$ to obtain the new global model $G^{t}$.
For direct comparison, we use the formula from~\cite{fedlearn_dp}:
\begin{equation}
    \label{eq:fa}
    G^{t} = G^{t-1} + \frac{\eta}{m}\sum_{i=1}^m (P_i^{t} - G^{t-1})
\end{equation}
All motivating applications of federated learning, such as predictive
keyboards and collaborative analysis of biomedical data, involve
participants with non-idd data, and federated learning is specifically
designed to accommodate training with millions of participants.  Recently,
a federated language model was trained on $7.5$ billion sentences from
1.5 million North American participants~\cite{hard2018federated}.



\paragraphbe{Adding privacy.}
ML models can leak their training data~\cite{song2017model,
shokri2017membership}.  In federated learning, participants' model
updates can leak even more~\cite{melis2018inference}.  Differential
privacy~\cite{dwork2008differential, dwork2011differential}
has been promoted as the solution to privacy problems in deep
learning~\cite{abadi2016deep} and federated learning~\cite{fedlearn_dp}.
Differential privacy (DP) provides $(\epsilon, \delta)$ privacy guarantee
when the federated mechanism $\mathcal{M}$ and two set of users $Q$, $Q'$
that differ by one participant produce models in any set $\mathcal{G}$
with probabilities that satisfy:
\begin{equation}
    \texttt{Pr}[\mathcal{M}(Q) \in \mathcal{G}] \leq e^{\epsilon} \;
    \texttt{Pr}[\mathcal{M}(Q') \in \mathcal{G}] + \delta    
\end{equation}


In practice, applying differential privacy to federated learning
involves (a) clipping each participant's update, and (b) adding random
noise~\cite{fedlearn_dp}.  Aggregation is modified as follows:
\begin{equation}
    \label{eq:dp}
    G^{t} = G^{t-1} + \frac{\eta}{m}\sum^m_{i=1}
    (\texttt{Clip}(P^{t}_i-G^{t-1}, S)) + \mathcal{N}(0,\sigma)        
\end{equation}
Achieving a given $(\epsilon,\delta)$ privacy guarantee involves carefully
selecting the clipping bound $S$ and noise $\sigma$ using the moments
accountant method~\cite{abadi2016deep}.  We omit the details and instead
using parameters from previous work with a similar setup~\cite{fedlearn_dp}.



\paragraphbe{Adding integrity.} 
Training with millions of participants is inherently vulnerable to
malicious participants who can prevent the training from converging
and/or inject a backdoor into the model~\cite{bagdasaryan2018backdoor}.
To ensure that malicious participants and other outliers cannot
influence the joint model, robust aggregation replaces average by
median~\cite{yin2018byzantine, chen2019distributed}:
\begin{equation}
    \label{eq:med}
    G^{t} = G^{t-1} + \eta(\tilde{P}^{t} - G^{t-1}),
\end{equation}
\noindent
where $\tilde{P}^{t}$ is the element-wise median among the updates
submitted in round $t$.  We focus on median aggregation, but our
adaptation techniques also apply to other so called ``Byzantine-tolerant''
aggregation schemes~\cite{blanchard2017machine, mhamdi2018hidden,
damaskinos2019aggregathor}.


\section{Tasks}
\label{sec:tasks}

We use two standard tasks from the federated learning literature:
next-word prediction and CIFAR-10 image classification~\cite{fedlearn_1}.
We evaluate the original averaging aggregation~\cite{fedlearn_1},
differentially private aggregation~\cite{fedlearn_dp}, and robust median
aggregation~\cite{chen2019distributed, yin2018byzantine}: BASIC-FED,
DP-FED, and ROBUST-FED, resp.


For DP-FED, we follow~\cite{fedlearn_dp} and use the clipping bound
$S=15$ and Gaussian noise with $\sigma=0.01$ for Equation~\ref{eq:dp}
(the model does not converge with bigger noise).  For ROBUST-FED, we
compute the coordinate-wise median instead of the mean of participants'
gradients.  Each participant trains locally using cross-entropy loss
$\mathcal{L}_{cross}(P,x)$.  All code was implemented in PyTorch 1.2
and executed on an Ubuntu 18.04 machine with 4 Nvidia GeForce RTX 2080
Ti GPUs and 12GB RAM.
We release our code publicly for
reproducibility.\footnote{\url{https://github.com/ebagdasa/federated_adaptation}}
\begin{figure*}[!htb]
\centering
\subfigure[BASIC-FED]{
    \begin{minipage}[t]{0.3\linewidth}
        \centering
        \includegraphics[width=1\linewidth]{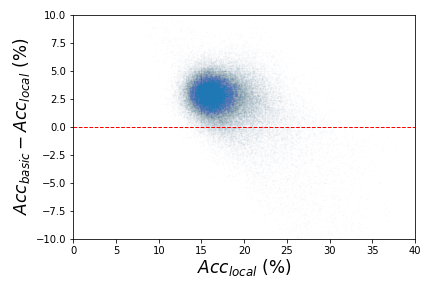}\\
        \includegraphics[width=0.96\linewidth]{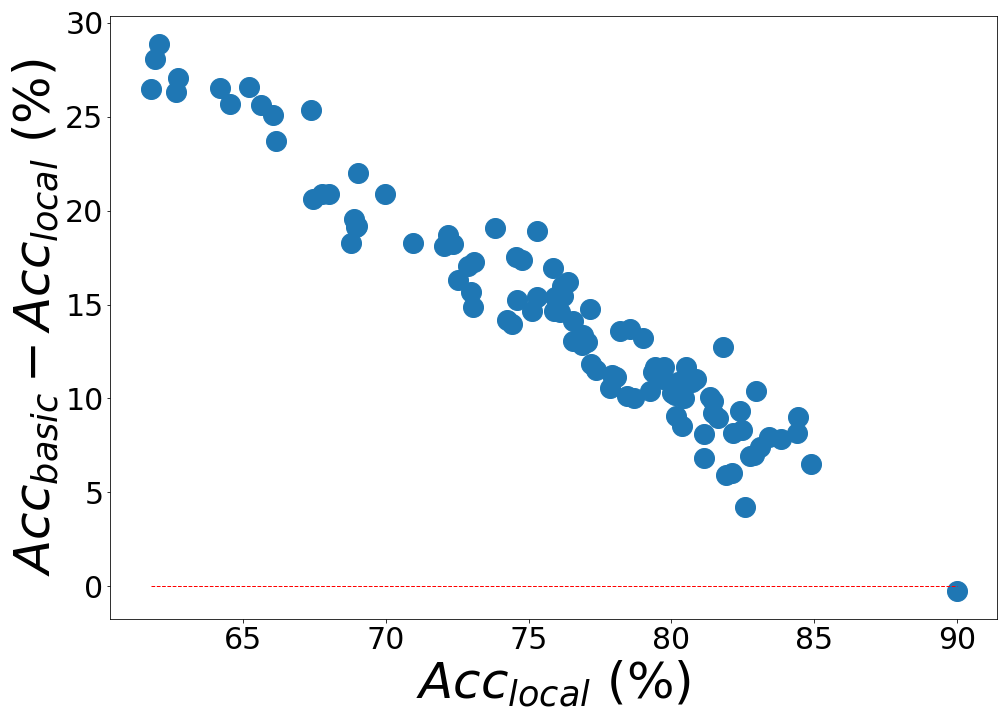}\\
    \end{minipage}%
}%
\subfigure[DP-FED]{
    \begin{minipage}[t]{0.3\linewidth}
        \centering
        \includegraphics[width=1\linewidth]{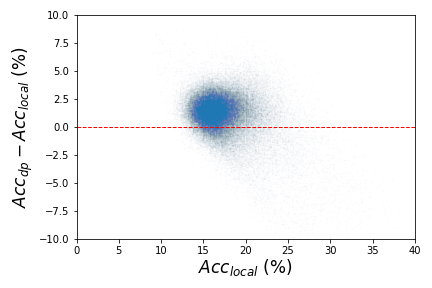}\\
        \includegraphics[width=0.96\linewidth]{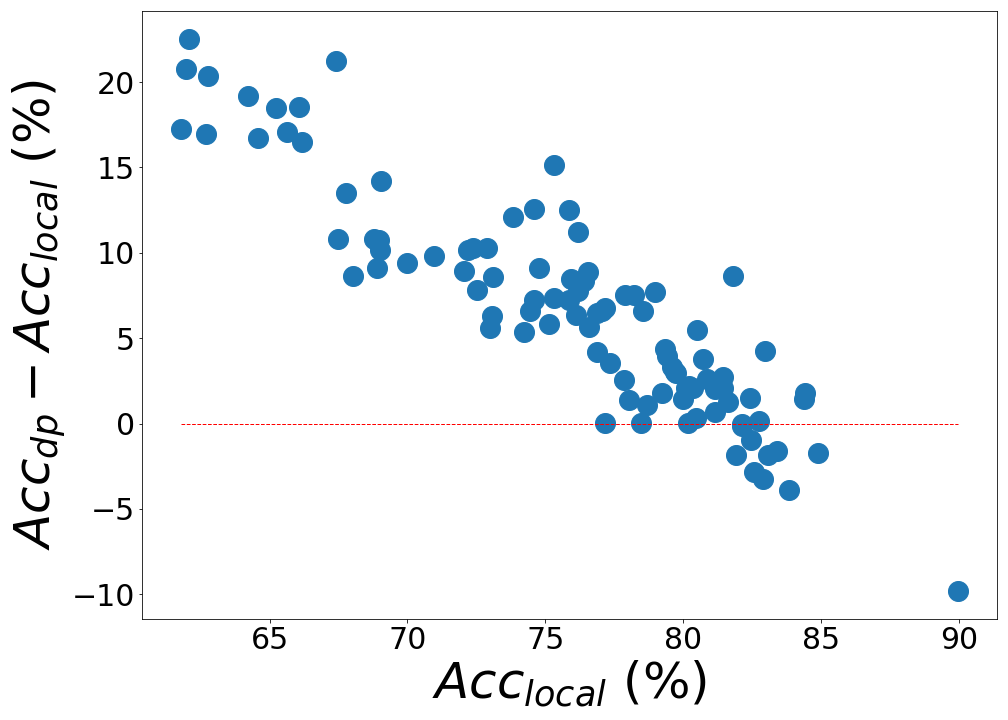}\\
    \end{minipage}%
}%
\subfigure[ROBUST-FED]{
    \begin{minipage}[t]{0.3\linewidth}
        \centering
        \includegraphics[width=1\linewidth]{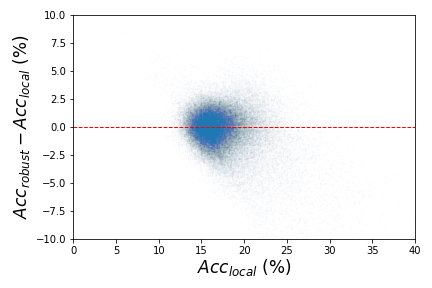}\\
        \includegraphics[width=0.96\linewidth]{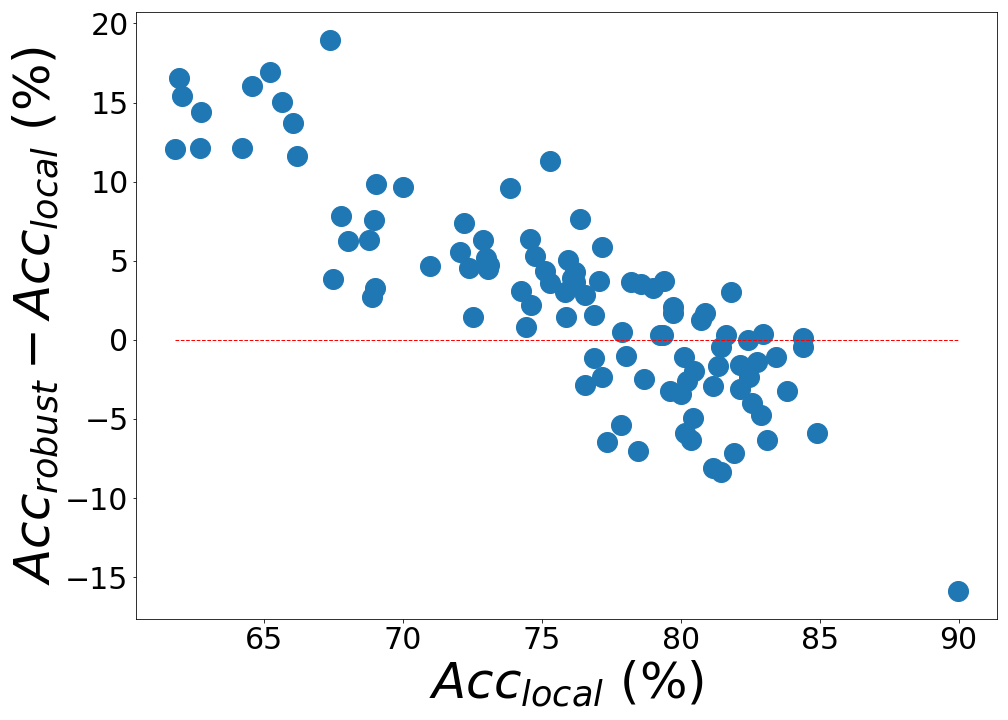}\\
    \end{minipage}%
}%
\centering
\caption{Accuracy improvements of federated models over local, trained-from-scratch models for word prediction (top row) and image classification (bottom row) tasks.}
\label{fig:fed_scr}
\end{figure*}
\subsection{Next-word prediction}

We train word prediction models on a randomly chosen month (November
2017) of the Reddit dataset~\cite{Reddit} with 80,000 participants
(i.e., Reddit users) who have between 150 and 500 posts, treating each
post as one sentence.  This task is a realistic application of federated
learning, involving unbalanced data from different distributions (see
Appendix~\ref{sec:imbalance} in supplementary materials).  Some users
have posts with a few simple, repeating phrases, while others write in
sophisticated prose.

We compiled a dictionary of $50,000$ most frequent words and replaced
all others with the \texttt{unk} token.  To create BASIC-FED,
DP-FED, and ROBUST-FED models, we train 2-layer LSTM models with
200 hidden units and 10 million parameters~\cite{pytorchwordmodel}.
Following~\cite{fedlearn_dp}, we train for 5,000 rounds with $m=100$
participants per round, aggregation learning rate $\eta=1$, batch size
20, and $B=2$ internal epochs using SGD.  For training participants'
models, we tried inner learning rates of 0.1, 1, 10, 20, and 40, yielding
global test accuracy of, respectively, 9.07$\%$, 14.34$\%$, 18.83$\%$,
19.20$\%$ and 19.29$\%$.  We thus set the inner learning rate to $lr=40$.
To measure test accuracy, we split each participant's Reddit posts into
the training and test sets in chronological order at the $9:1$ ratio.

\subsection{Image classification}

We split the CIFAR-10~\cite{krizhevsky2009learning} training set into
100 participants.  To simulate a non-iid distribution, we allocate
images from each class to participants using Dirichlet distribution
with $\alpha=0.9$, similar to~\cite{hsu2019measuring}.  We train all
federated models for 1,000 rounds with the aggregation learning rate
$\eta=1$ and batch size of 32.  Following~\cite{fedlearn_1}, in every
round we aggregate 10 randomly selected participants, each of whom trains
a ResNet-18 model (with 11.2 million parameters) with the inner learning
rate of $0.1$ and $B=2$ internal epochs using SGD with momentum $0.9$
and weight decay $0.0005$.

CIFAR-10 is not divided into distinct participants.  To measure the test
accuracy of a model on a participant's distribution, we calculate its
per-class accuracy on the CIFAR-10 global test dataset, multiply it by
the corresponding class's ratio in the participant's training dataset,
and sum up the resulting values.

\section{Privacy and robustness destroy individual accuracy}
\label{sec:fedworselocal}
\begin{figure*}[!htb]
\centering
\subfigure[BASIC-FED]{
    \begin{minipage}[t]{0.3\linewidth}
        \centering
        \includegraphics[width=1\linewidth]{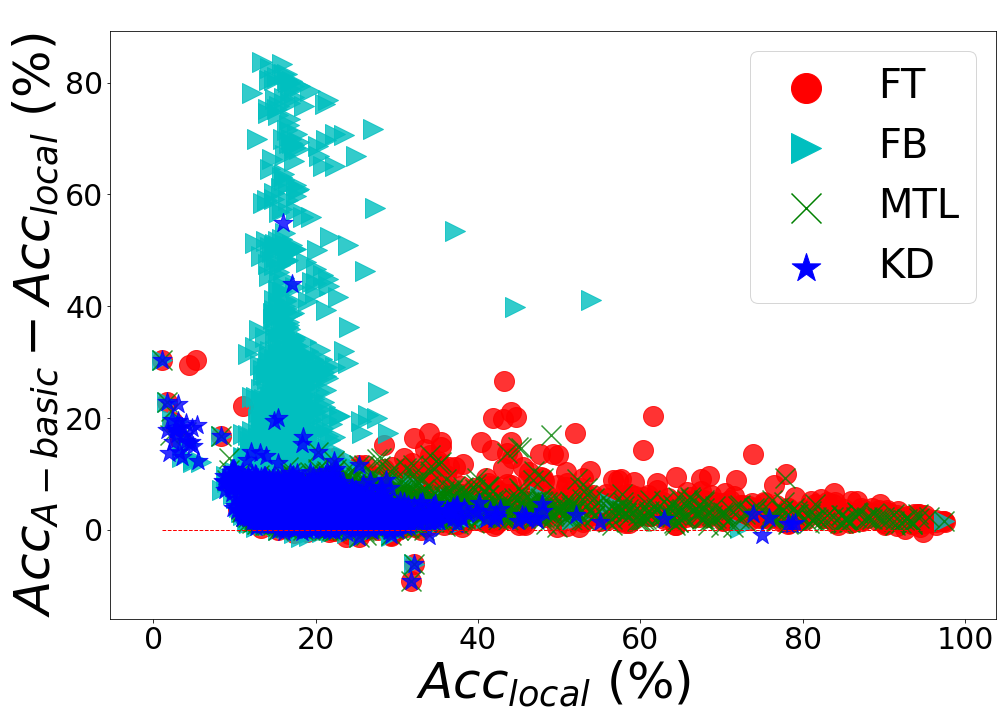}\\
        \includegraphics[width=1\linewidth]{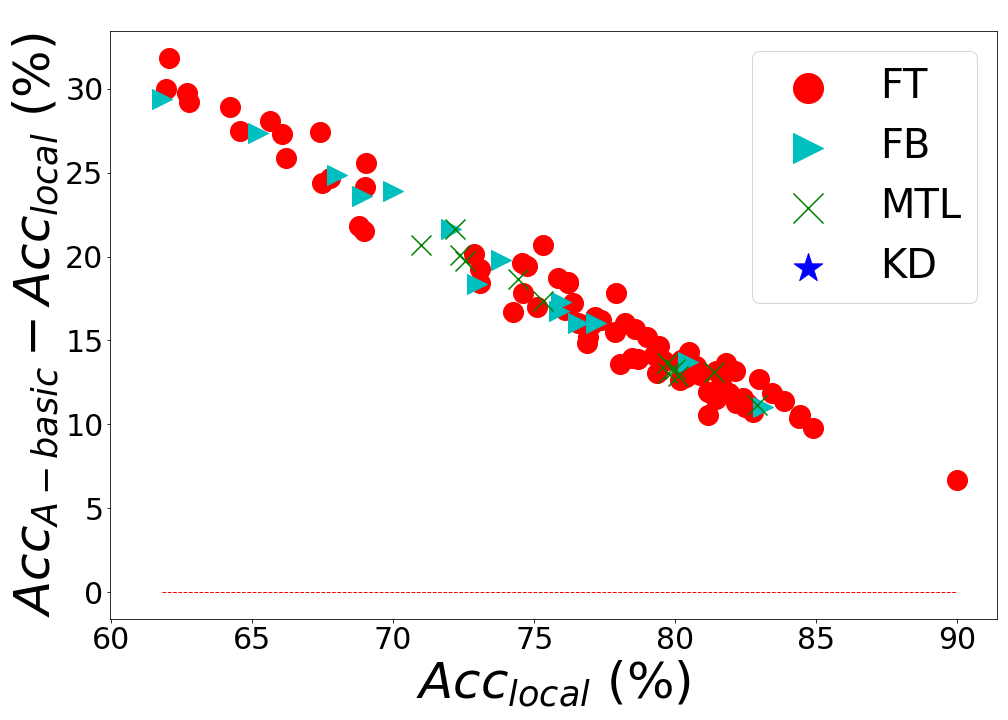}\\
    \end{minipage}%
}%
\subfigure[DP-FED]{
    \begin{minipage}[t]{0.3\linewidth}
        \centering
        \includegraphics[width=1\linewidth]{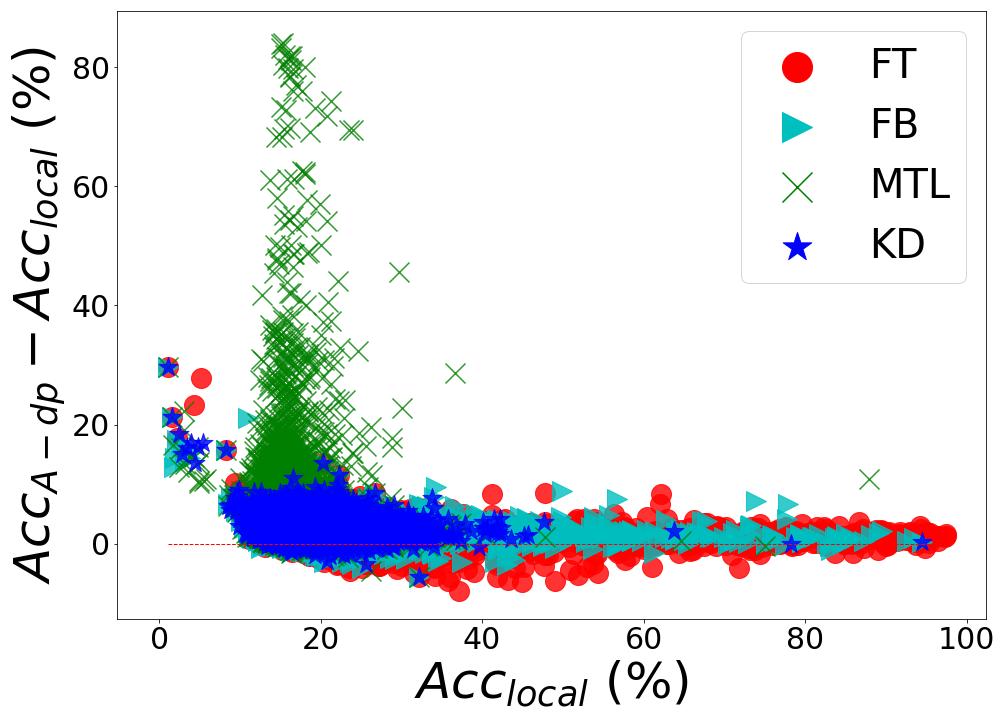}\\
        \includegraphics[width=1\linewidth]{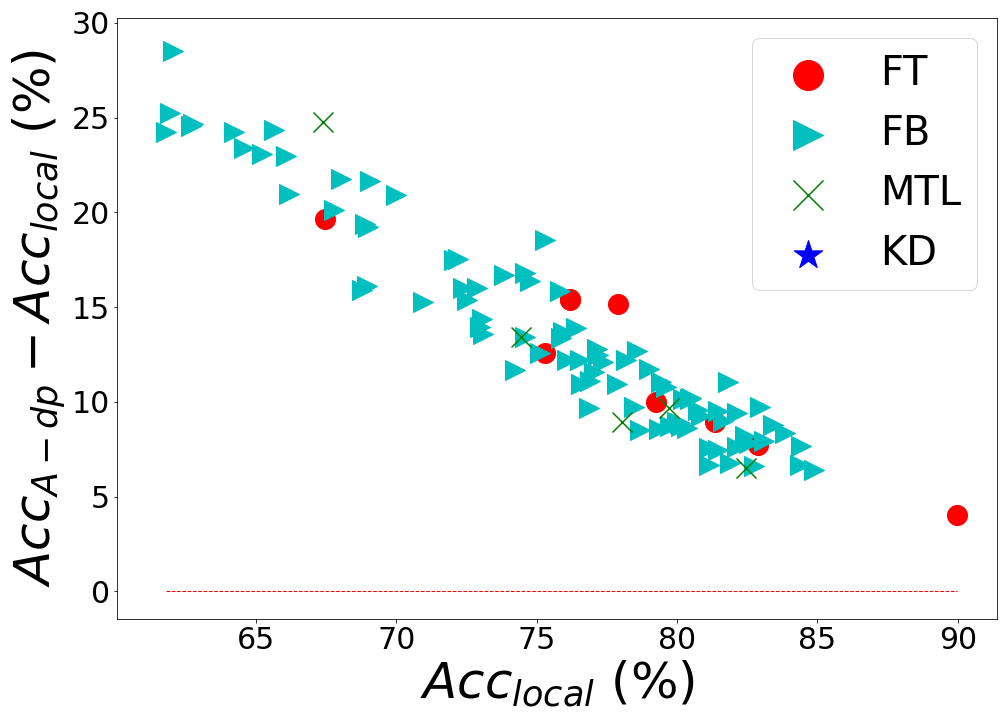}\\
    \end{minipage}%
}%
\subfigure[ROBUST-FED]{
    \begin{minipage}[t]{0.3\linewidth}
        \centering
        \includegraphics[width=1\linewidth]{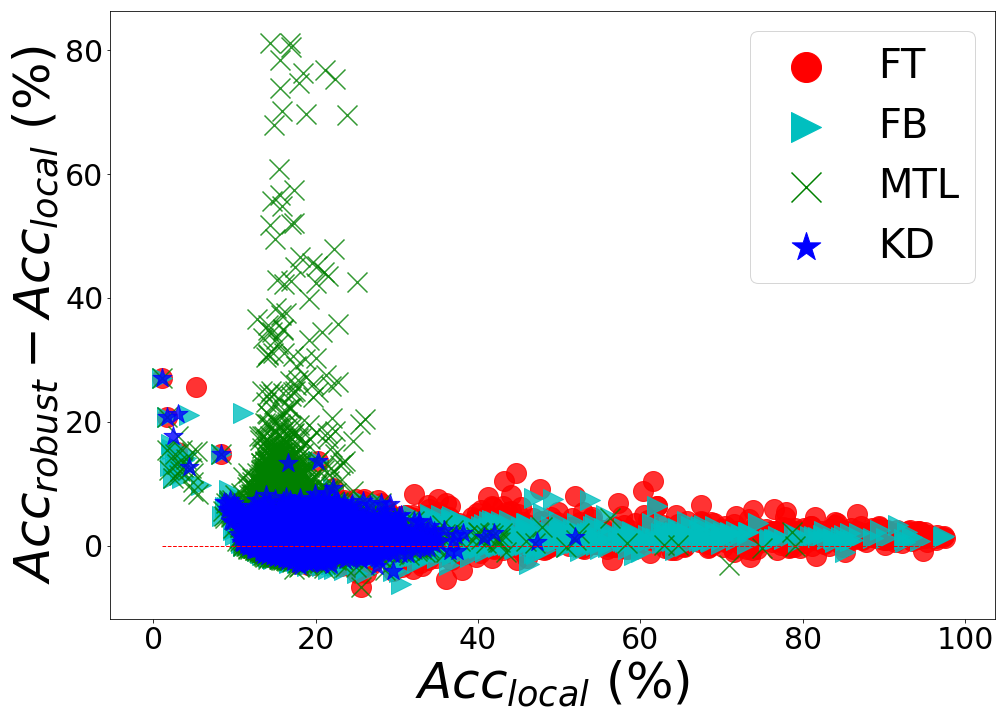}\\
        \includegraphics[width=1\linewidth]{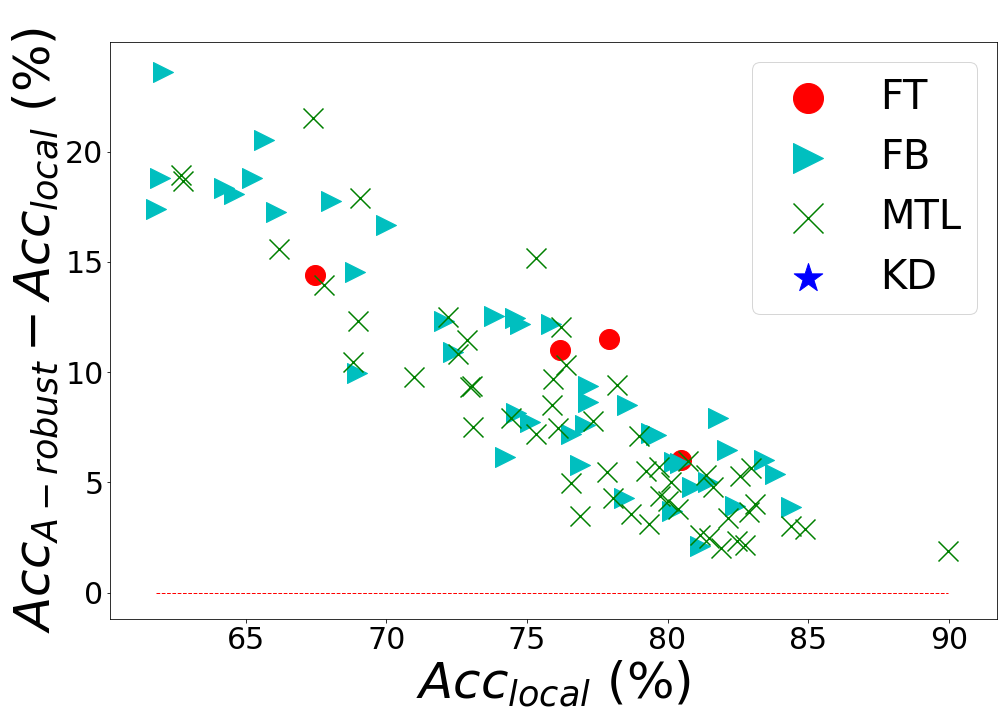}\\
    \end{minipage}%
}%
\centering
\caption{Accuracy improvements of adapted federated models over local, trained-from-scratch models for word prediction (top row) and image classification (bottom row) tasks.}
\label{fig:adapted_fed_scr}
\end{figure*}

Federated learning relies on the participation of thousands or millions
of users.  Some may be motivated by altruism, but rational users need an
incentive to participate.  For example, they benefit if the federated
model is more accurate than the models they can train locally on their
own data.

Accuracy of federated models is typically measured\textemdash often
on tasks such as MNIST that are not representative of the intended
applications of federated learning\textemdash on a holdout dataset
compiled from all participants' data~\cite{fedlearn_1}.  When the
participants are not iid and some have their own, idiosyncratic data,
global accuracy does not represent whether the model is accurate for a
specific participant.

In realistic motivating scenarios, such as predictive keyboards, fraud
detection, and biomedical research, an aggregated model may perform
well on the global test data but poorly on an individual participant's
test data, thus removing their main incentive to participate.  In the
rest of this paper, we focus on the accuracy of federated models for
individual participants.



\autoref{fig:fed_scr}(a) compares the BASIC-FED models with the local,
trained-from-scratch models of individual participants.  The BASIC-FED
word prediction model (top row) is worse than the local models of 9.22$\%$
(7377) participants, which are trained for 100 epochs with the learning
rate of 1.  On the image classification task (bottom row), there is less
diversity among participants and BASIC-FED outperforms the local models
of all but 1 participant.  The local models were trained for 500 epochs
with the learning rate of 0.001.


With privacy or integrity protections, the comparison is unfavorable for
federated learning.  Figure~\ref{fig:fed_scr}(b) shows that DP-FED is less
accurate than the local models of many participants: 16,931 (21.16$\%$)
on word prediction (top row) and 11 (11$\%$) on image classification
(bottom row).  Even worse, ROBUST-FED shown in Figure~\ref{fig:fed_scr}(c)
is \textbf{less accurate than the local models for the majority of
participants} (41720, or 52.15$\%$) on word prediction and 34 (34$\%$)
on image classification.


These results illustrate the \textbf{tradeoff at the heart of federated
learning}.  To learn a joint model that is accurate for individual
participants, aggregation must incorporate contributions from every
participant. To protect data privacy and model integrity, aggregation must
limit the influence of these contributions, producing an inaccurate model.


\section{Local adaptation}
\label{sec:adaptation}
\begin{figure*}[!htb]
\centering
\subfigure[BASIC-FED]{
    \begin{minipage}[t]{0.3\linewidth}
        \centering
        \includegraphics[width=1\linewidth]{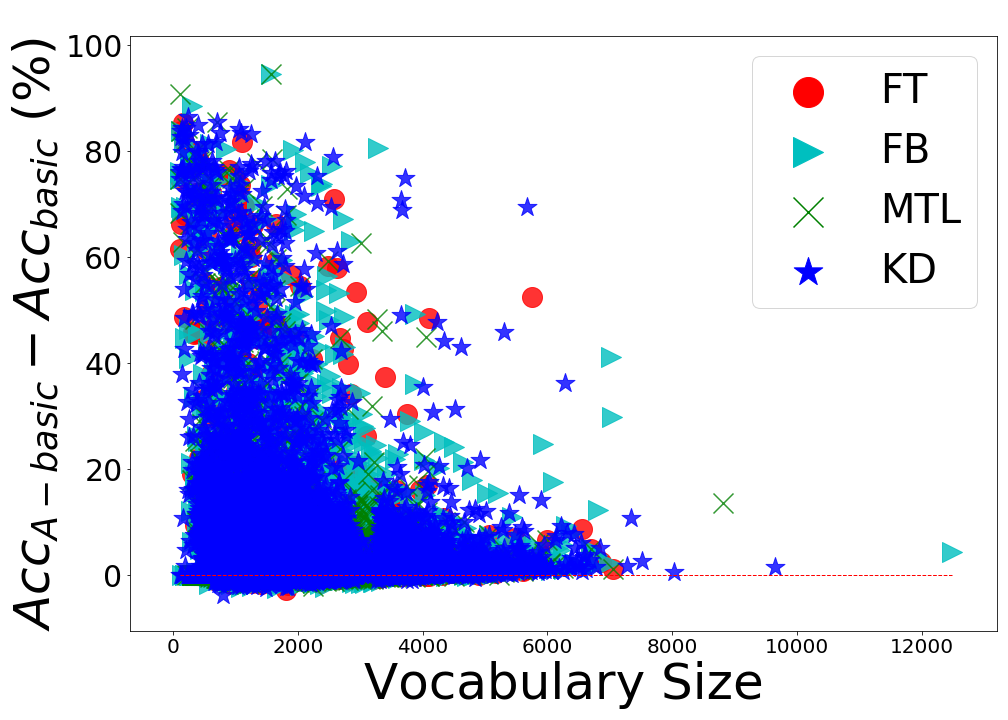}\\
        \vspace{0.02cm}
        \includegraphics[width=1\linewidth]{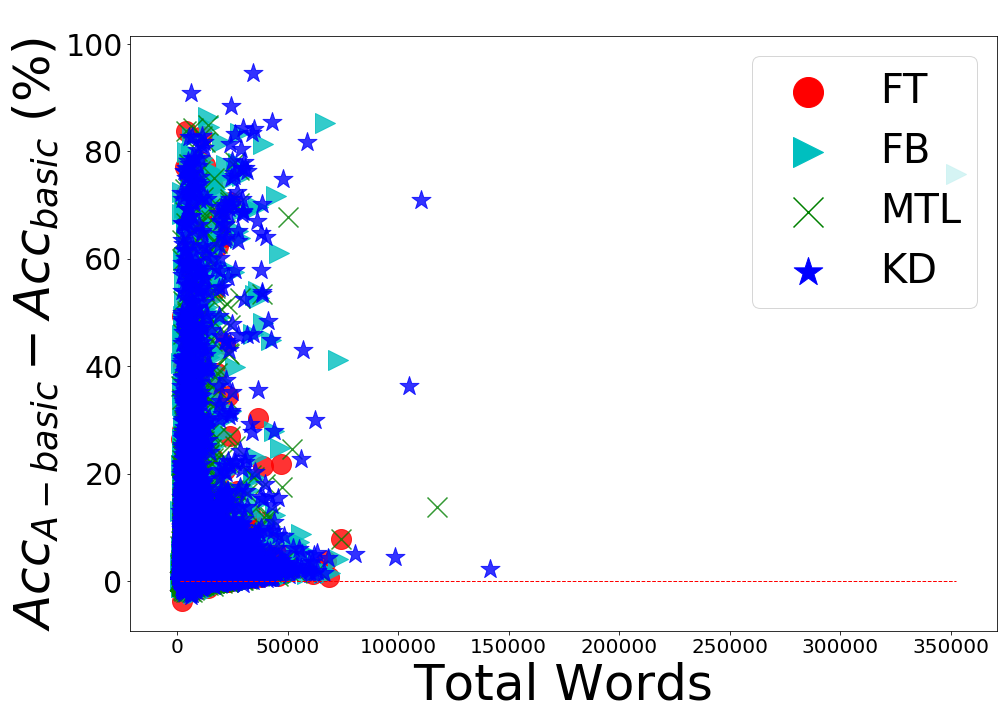}\\
        \vspace{0.02cm}
    \end{minipage}%
}%
\subfigure[DP-FED]{
    \begin{minipage}[t]{0.3\linewidth}
        \centering
        \includegraphics[width=0.97\linewidth]{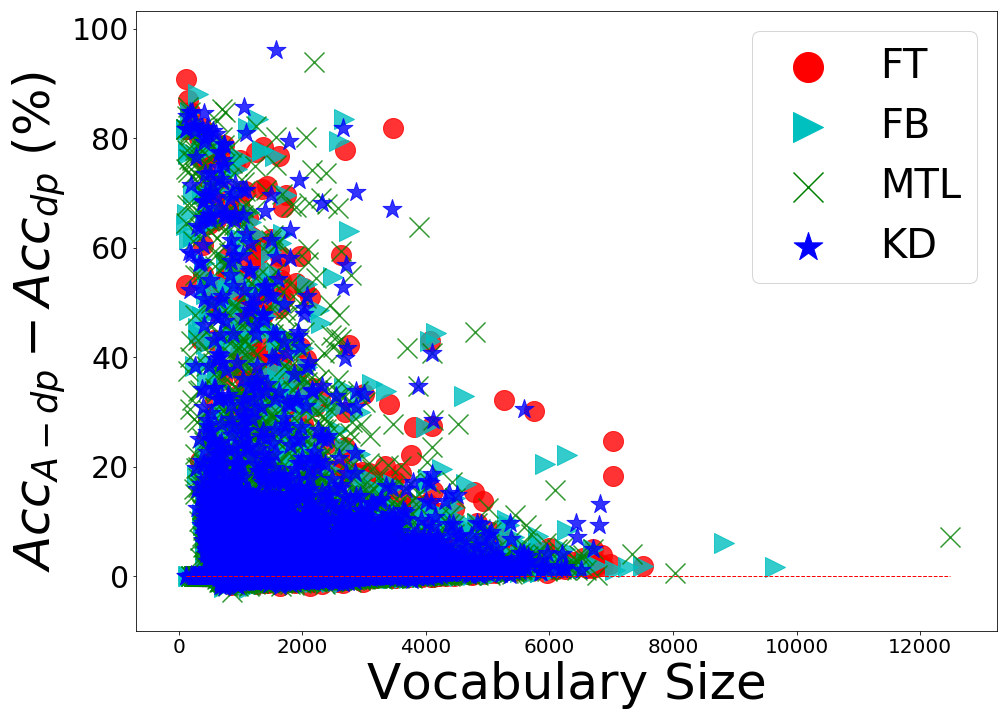}\\
        \vspace{0.2cm}
        \includegraphics[width=0.97\linewidth]{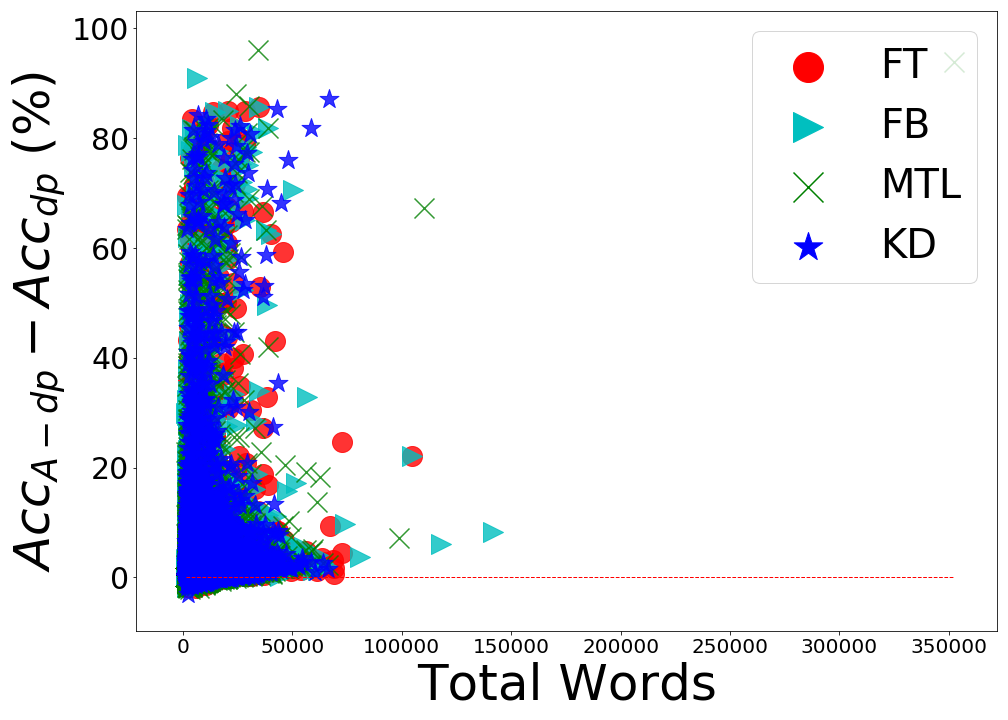}\\
        \vspace{0.02cm}
    \end{minipage}%
}%
\subfigure[ROBUST-FED]{
    \begin{minipage}[t]{0.3\linewidth}
        \centering
        \includegraphics[width=1.04\linewidth]{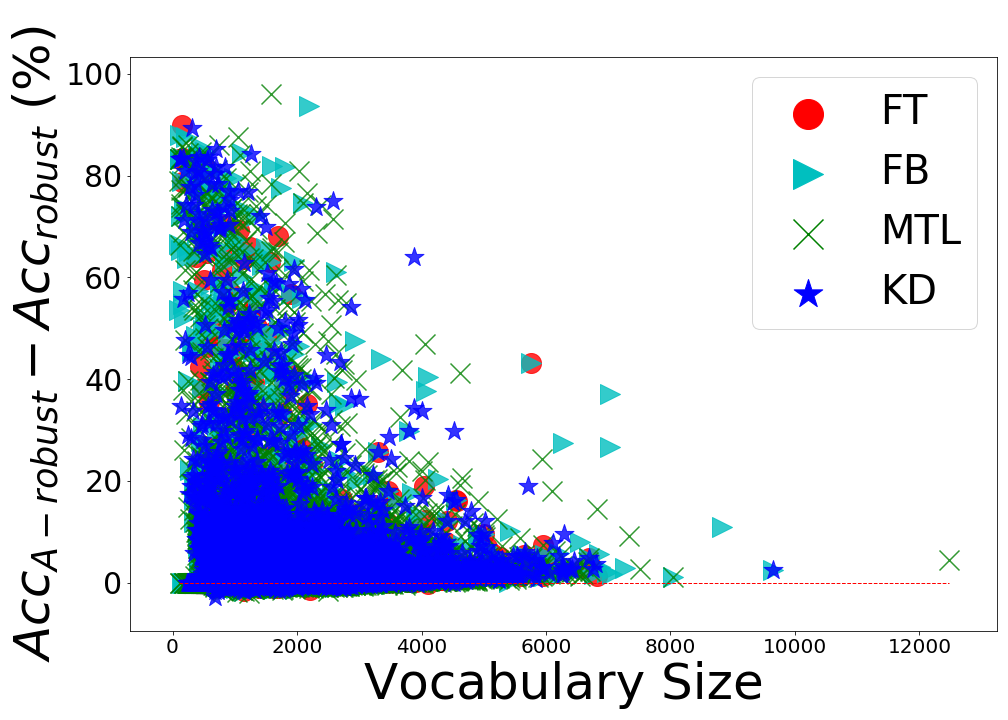}\\
        \vspace{0.02cm}
        \includegraphics[width=1.04\linewidth]{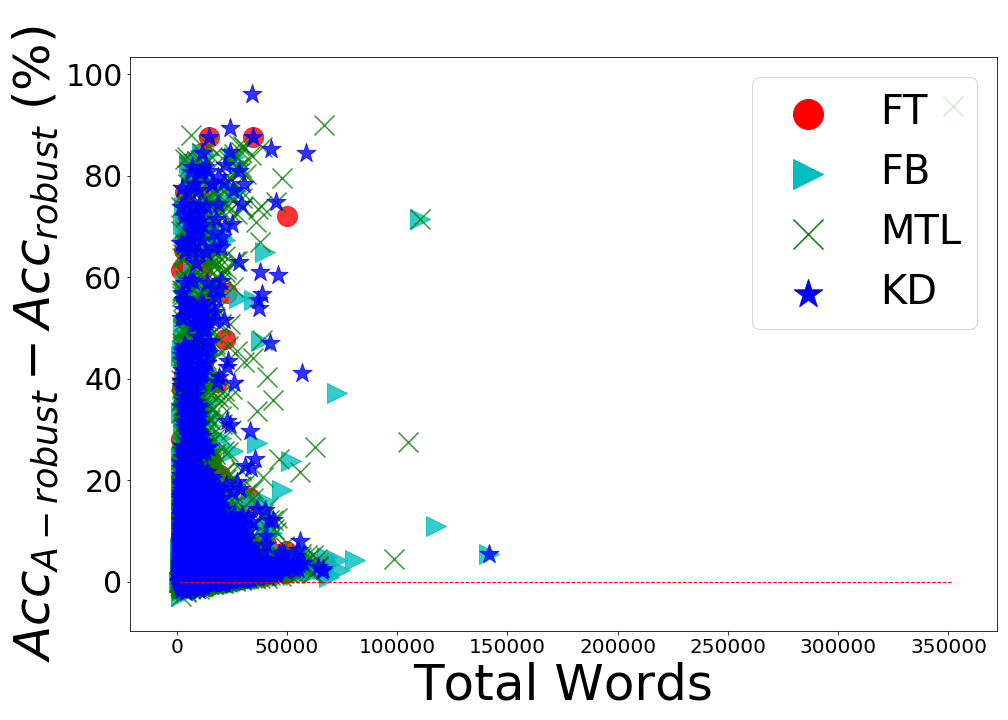}\\
        \vspace{0.02cm}
    \end{minipage}%
}%
\centering
\caption{Accuracy improvements of adapted over unadapted federated
models vs.\ vocabulary size (top row) and total words (bottom row).}
\label{fig:adapt_nonadapt_voc_total}
\end{figure*}

We investigate several techniques for adapting the federated model to
an individual participant.  Local word prediction models are trained for
$B=100$ epochs with the learning rate of 1, image classification models
for $B=200$ epochs with the learning rate of 0.001.

\paragraphbe{Fine-tuning (FT).}
Fine-tuning is a natural adaptation technique, used, e.g.,
in~\cite{wang2019federated}.  It re-trains all parameters of a trained
federated model on the participant's local training data (using the
above hyperparameters).  Fine-tuning takes advantage of the federated
model's feature extraction network instead of learning it from scratch.
\emph{\textbf{Freeze-base (FB)}} is a variant that freezes the base layers
of the federated model and fine-tunes only the top layer.  When using
fine-tuning for local adaptation, we experimented on 1,000 participants
with the learning rates of 0.1, 1, and 10, yielding mean accuracy of,
respectively, 20.58$\%$, 20.99$\%$ and 18.28$\%$.  Therefore, we set
$lr=1$.

\paragraphbe{Multi-task learning (MTL).}
With non-iid distribution, a participant's local data may be very
different from the other participants.  To mitigate overfitting, we
treat local adaptation as a multi-task learning problem, where task
$X$ requires high performance on the union of all participants and
task $Y$ requires high performance on a single participant.  We take
the federated model $G^T$ optimized for task $X$ and aim to create
an adapted model $A$ (initialized as $G^T$) optimized for task $Y$.
To overcome the catastrophic forgetting~\cite{french1999catastrophic}
of task $X$ while learning $Y$, we use elastic weight
consolidation~\cite{kirkpatrick2017overcoming} which selectively
slows down learning on the weights important for $X$.  To learn task
$Y$, we use the same cross-entropy loss $\mathcal{L}_{cross}$ as in
Section~\ref{sec:tasks} and aim to minimize:
\begin{equation} \label{eq:mtl}
    \ell(A, x)=\mathcal{L}_{cross}(A, x)+\sum\limits_i\frac{\lambda}{2}F_i(A_i-G^T_i)^2    
\end{equation}
where $\lambda$ is the importance of task $X$ vs.\ $Y$, $F$ is
the Fisher information matrix (computed on a public auxiliary dataset), $i$ is the label of each parameter.
Following~\cite{kirkpatrick2017overcoming}, we use $\lambda=5000$.


\paragraphbe{Knowledge distillation (KD).}
Knowledge distillation~\cite{hinton2015distilling} extracts information
from a ``teacher'' model into a ``student'' model.  We treat the federated
model $G^T$ as the teacher and the adapted model $A$ as the student,
except that in our case both models have the same structure and $A$ is
initialized to $G^T$ but the local dataset on which $A$ is trained is
a small subset of the dataset on which $G^T$ is trained.  We conjecture
that enforcing the similarity of logits between $G^T$ and $A$ using the
loss function from the knowledge distillation literature helps mitigate
overfitting on the small dataset.

We represent $G^T(x), A(x)$ as the pre-softmax logit outputs of the two
models and minimize:
\begin{equation} 
\label{eq:kd}
\begin{split}
    \ell(A, x)
    &
    =\alpha K^2\mathcal{L}_{cross}(A, x)   \\
    &
    +  (1-\alpha)\textsc{KL}(\sigma(\frac{G^T(x)}{K}), \sigma(\frac{A(x)}{K}))
\end{split}
\end{equation}
\noindent
where $\textsc{KL}$ is Kullback-Leibler divergence loss, $\sigma$
is softmax, $\alpha$ is the weight parameter, $K$ is the temperature
constant.  The $K^2$ term equalizes gradient magnitudes for both losses.
We did not observe significant differences when varying $\alpha=[0.1,
0.5, 0.95, 0.99]$ and $K=[4,6,10]$ and set $\alpha=0.95, K=6$.


\section{Local adaptation gives an incentive to participate in federated learning}
\label{sec:experiments}

We investigate the effects of the FT, FB, MTL, KD adaptation techniques
from Section~\ref{sec:adaptation} on the accuracy of BASIC-FED, DP-FED,
ROBUST-FED models for individual participants.  Dots and bars in the
figures are color-coded according to the technique that yielded the best
accuracy improvement.

For the word-prediction task, there are 80,000 participants, but we only
adapt the 79,097 participants whose vocabulary size (i.e., number of
unique symbols) is over 100, the percentage of utility symbols (e.g.,
punctuation) is under 40$\%$, and the difference between total and
utility symbols is over 1,000.

\begin{figure*}[!htb]
\centering
\subfigure[BASIC-FED]{
    \begin{minipage}[t]{0.3\linewidth}
        \centering
        \includegraphics[width=1\linewidth]{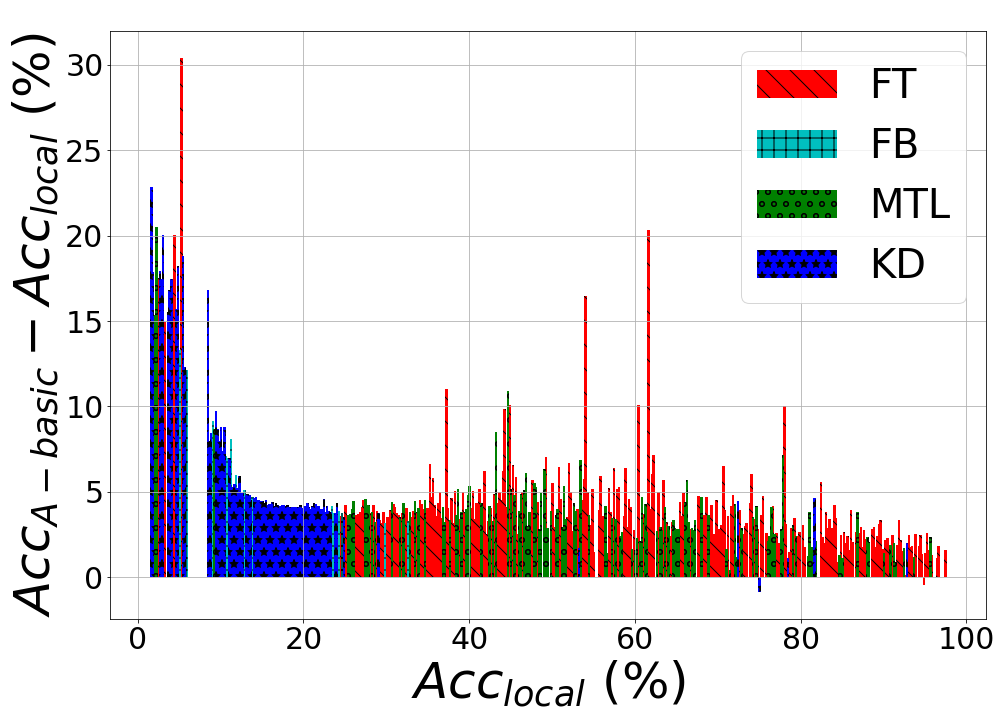}\\
        \vspace{0.02cm}
        \includegraphics[width=1\linewidth]{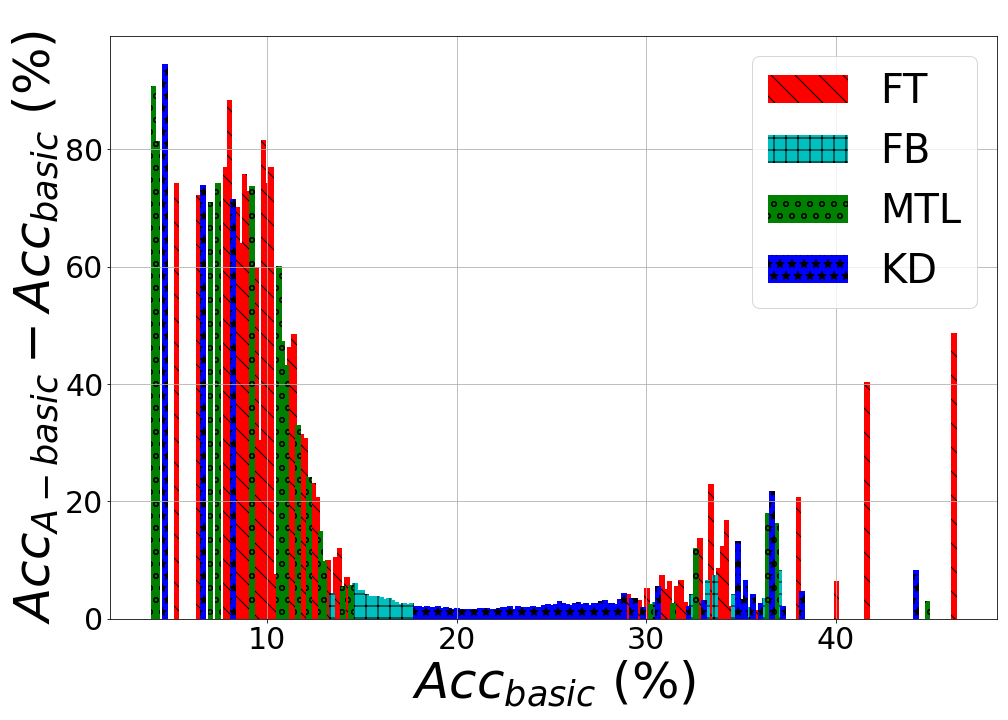}\\
        \vspace{0.02cm}
    \end{minipage}%
}%
\subfigure[DP-FED]{
    \begin{minipage}[t]{0.3\linewidth}
        \centering
        \includegraphics[width=0.97\linewidth]{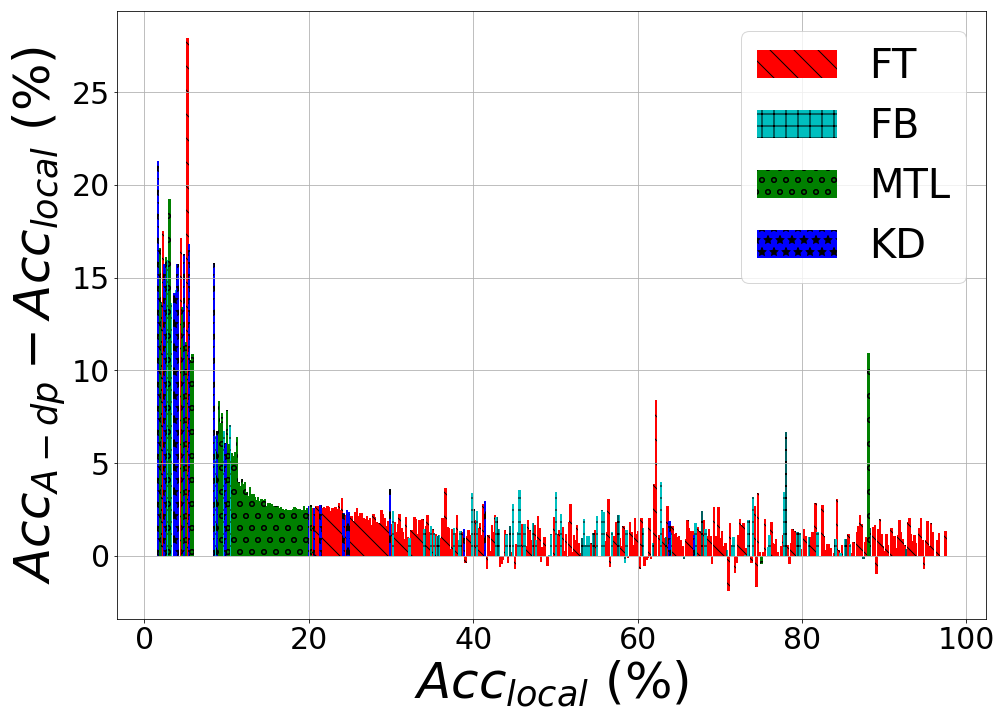}\\
        \vspace{0.2cm}
        \includegraphics[width=0.97\linewidth]{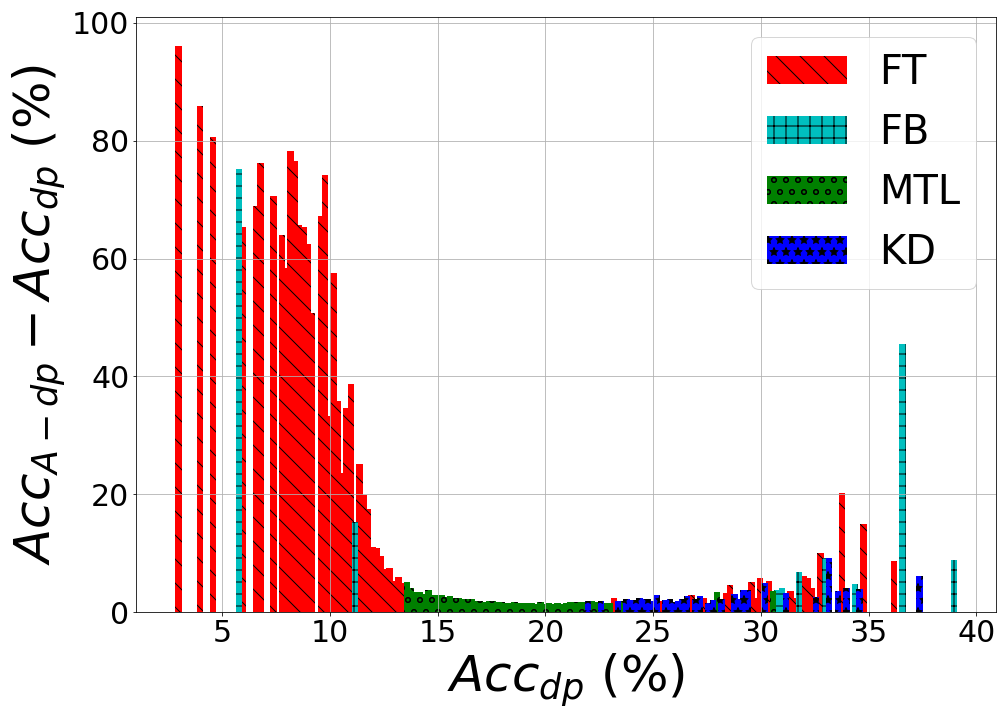}\\
        \vspace{0.02cm}
    \end{minipage}%
}%
\subfigure[ROBUST-FED]{
    \begin{minipage}[t]{0.3\linewidth}
        \centering
        \includegraphics[width=1.01\linewidth]{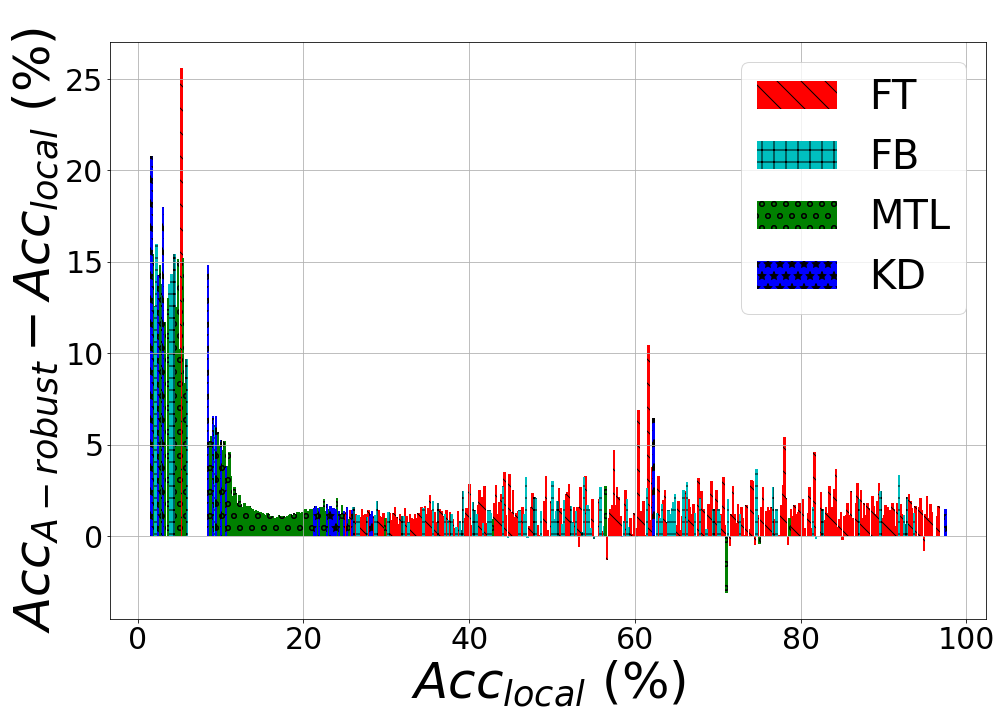}\\
        \vspace{0.02cm}
        \includegraphics[width=1.01\linewidth]{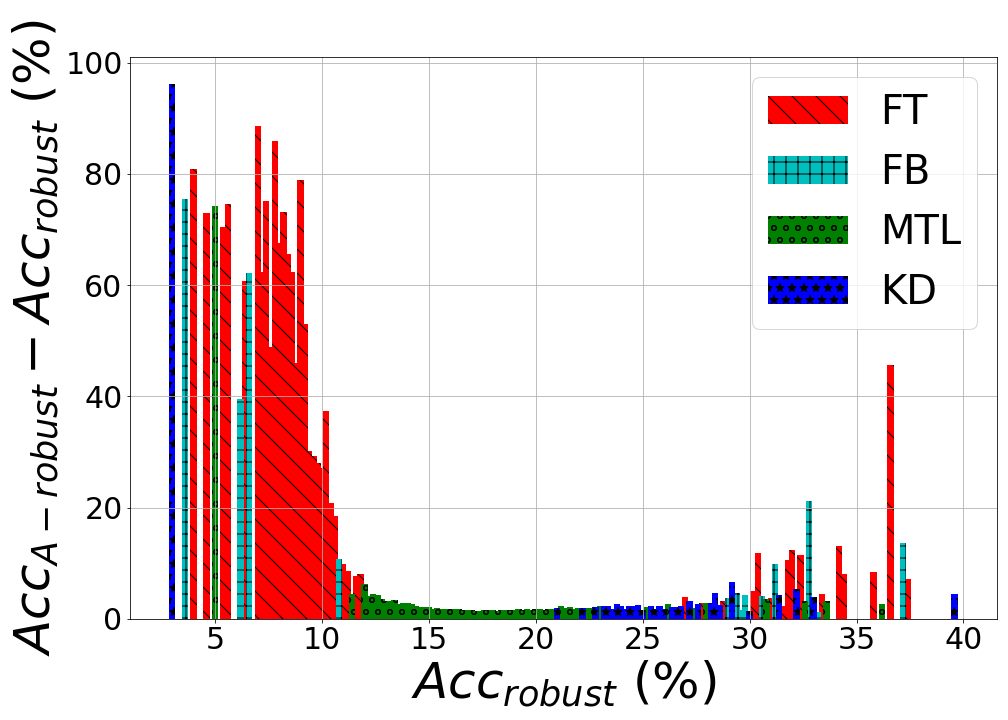}\\
        \vspace{0.02cm}
    \end{minipage}%
}%
\centering
\caption{Accuracy improvements of adapted federated models over
local (top row) and unadapted federated models (bottom row).}
\label{fig:adapt_non_scratch}
\end{figure*}

\subsection{Results of adaptation}
\label{subsec:adaptresults}

On word prediction, mean accuracy improvements due to adaptation are
2.32$\%$, 2.12$\%$, and 2.12$\%$ for BASIC-FED, DP-FED and ROBUST-FED,
respectively.  These improvements make up the loss of accuracy due
to differential privacy (-1.42\%) and robust aggregation (-2.81\%).
On image classification, mean accuracy improvements due to adaptation are
2.98$\%$, 6.83$\%$, and 6.34$\%$ for BASIC-FED, DP-FED, and ROBUST-FED,
respectively.  These improvements make up the loss of accuracy due to
differential privacy (-7.83$\%$) and robust aggregation (-11.89$\%$).

\autoref{fig:adapted_fed_scr}(a) shows the improvements of the adapted
BASIC-FED model over the participants' local models for word prediction
in the top row.  There are only 28 (0.04$\%$) participants for whom
the adapted BASIC-FED underperforms the local model.  
median accuracy of the adapted BASIC-FED are 22.41$\%$ and 21.00$\%$.
The bottom row of \autoref{fig:adapted_fed_scr}(a) shows the results for
image classification.  Adapted BASIC-FED outperforms all local models.




\autoref{fig:adapted_fed_scr}(b) shows the improvements of the adapted
DP-FED model over the participants' local models for word prediction
in the top row.  There are only 1465 (1.85$\%$) participants for
whom the adapted DP-FED underperforms the local model.  The bottom
row of \autoref{fig:adapted_fed_scr}(b) shows the results for image
classification.  Adapted DP-FED outperforms all local models.




\autoref{fig:adapted_fed_scr}(c) shows the improvements of the adapted
ROBUST-FED model over the participants' local models for word prediction
in the top row.  There are 14809 (18.72$\%$) participants for whom
the adapted ROBUST-FED underperforms the local model.  The bottom
row of \autoref{fig:adapted_fed_scr}(c) shows the results for image
classification.  Adapted ROBUST-FED outperforms all local models.



\subsection{Analysis}
\label{subsec:adaptionanalysis}

Our baselines are respective accuracies of (1) the participant's local
model and (2) the unadapted federated model, both measured on that
participant's test data.

\paragraphbe{Adapted models vs.\ trained-from-scratch models.}
In \autoref{subsec:adaptresults}, we showed that the adapted federated
models outperform the local models for most participants.  Top row of
\autoref{fig:adapt_non_scratch} visualizes the effects of adaptation
on different types of participants.  Accuracy is divided into $0.2\%$
intervals and the improvements for all participants whose local model
accuracy falls into a given interval are averaged, yielding a single bar.
The color of the bar corresponds to the adaptation technique that accounts
for the biggest share of the total improvement of the participants in
the interval.

Participants with inaccurate local models are on the left side of
the X-axis.  The original federated model was already more accurate
than their local models (\autoref{fig:fed_scr}), yet local adaptation
yields the biggest improvements for them and thus a stronger incentive
to participate.

Participants with accurate local models did not benefit from federated
learning (\autoref{fig:fed_scr}(a)), but adaptation now gives them an
incentive to participate because the adapted model outperforms the local
model\textemdash even though the improvement is smaller than for the
low-accuracy participants.
\begin{figure*}[!htb]
\begin{tabular}{cc}
\begin{minipage}[t]{0.33\linewidth}
    \includegraphics[width = 1\linewidth]{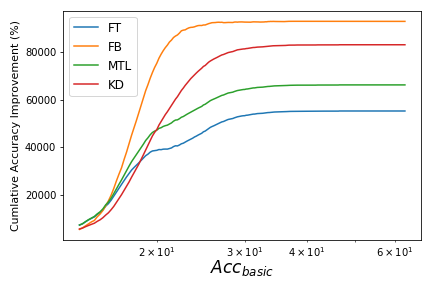}
\end{minipage}
\begin{minipage}[t]{0.33\linewidth}
    \includegraphics[width = 1\linewidth]{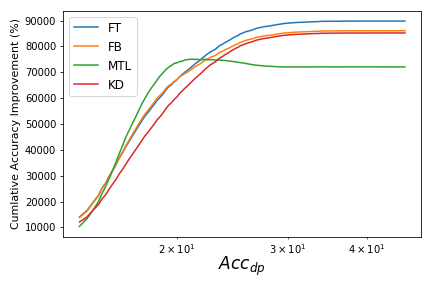}
\end{minipage}
\begin{minipage}[t]{0.33\linewidth}
    \includegraphics[width = 1\linewidth]{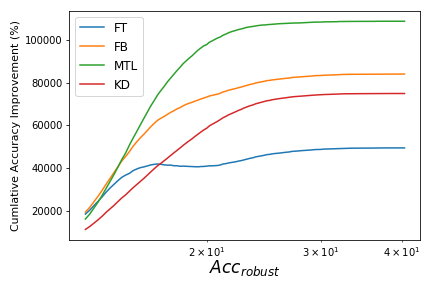}
\end{minipage}
\end{tabular}
\caption{Cumulative accuracy improvements of different adaptations on BASIC-FED (left), DP-FED (middle), and ROBUST-FED (right).}
\label{fig:text_cai}
\end{figure*}

\paragraphbe{Adapted vs.\ unadapted federated models.}
Bottom row of \autoref{fig:adapt_non_scratch} shows how adaptation
improves the accuracy of federated models.  The biggest improvements
accrue to ``tail'' participants whose local models have low accuracy.
Adaptation also improves the federated model for the ``head''
participants, for whom the unadapted model is already accurate.

\begin{figure}[!htb]
\centering
\begin{tabular}{cc}
\hspace{-1.5em}
\begin{minipage}[t]{0.51\linewidth}
    \includegraphics[width = 1.0\linewidth]{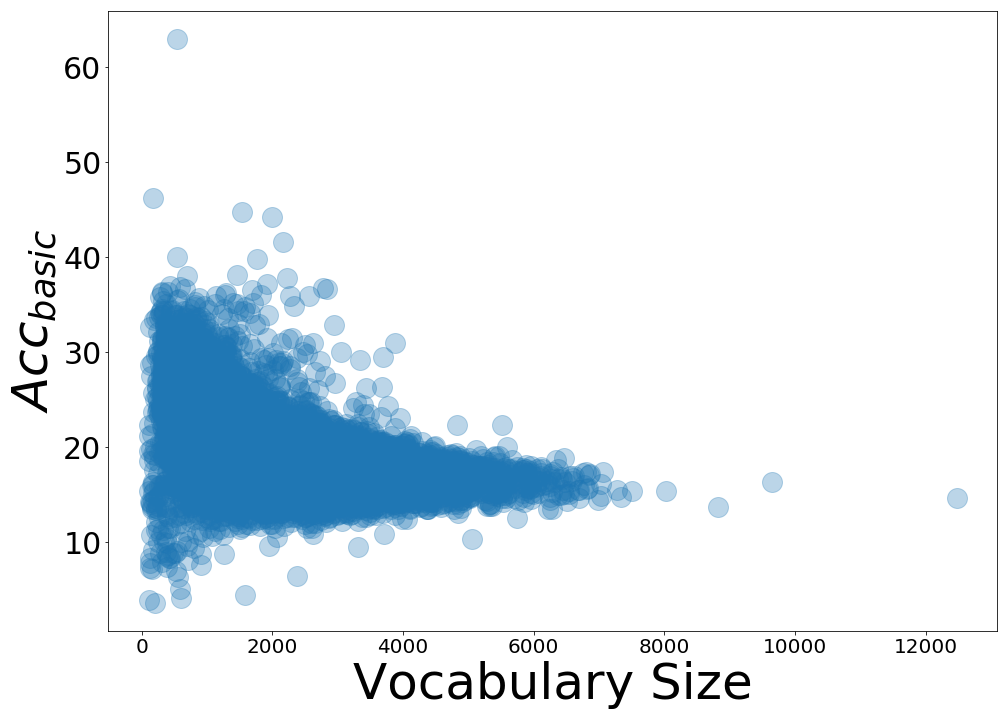}
\end{minipage}
\begin{minipage}[t]{0.51\linewidth}
    \includegraphics[width = 1.0\linewidth]{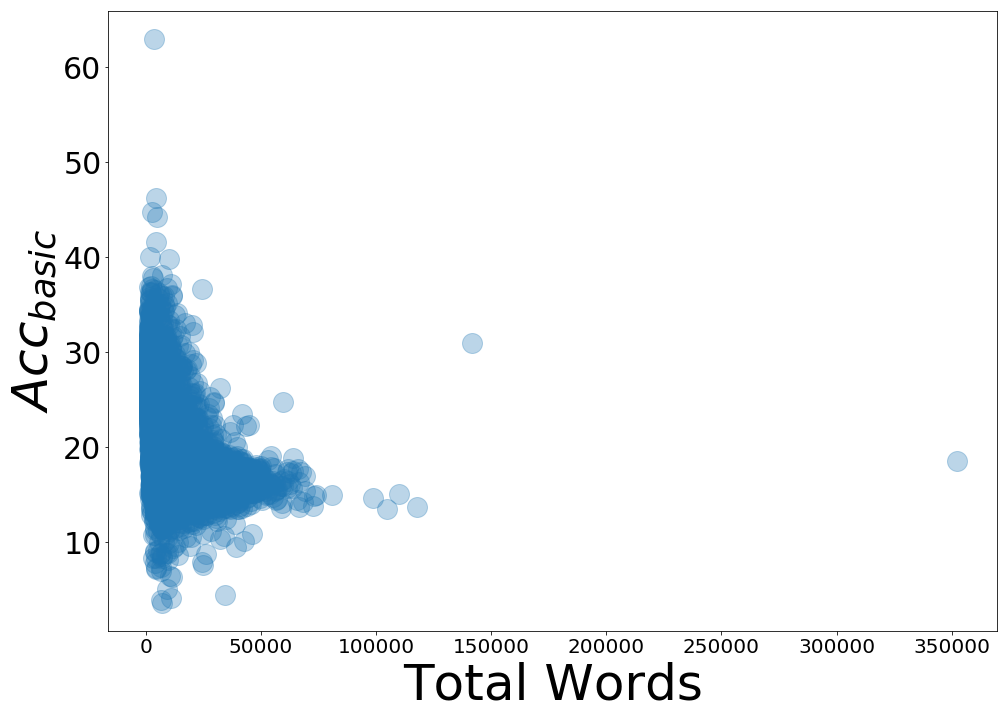}
\end{minipage}
\end{tabular}
\caption{Accuracy of the BASIC-FED model vs.\ vocabulary size (left) and total
words (right).}
\label{fig:basic_voc_total}
\end{figure}

To explain these effects, we measure the size (total number of words) and
complexity (vocabulary, i.e., number of unique words) for each participant
in our Reddit-based corpus.  \autoref{fig:adapt_nonadapt_voc_total}
plots accuracy improvement vs.\ these features.  Adaptations
improve accuracy the most for the participants with simple (small
vocabulary) and small (few total words) data.  We conjecture that the
participants who obtain large accuracy improvements in the bottom row
of \autoref{fig:adapt_non_scratch} have simpler, smaller data.  To show
this for the BASIC-FED model, \autoref{fig:basic_voc_total} plots the
relationship between model accuracy and vocabulary size (respectively,
total words).

The participants with the highest \emph{and} lowest BASIC-FED accuracy
indeed have few, simple words.  We hypothesize that ``tail'' participants
(i.e., those with low BASIC-FED accuracy) use regular sentences that are
similar to other participants: e.g., `appreciation series has posts for
an author you mentioned.'  The low accuracy of BASIC-FED is simply due
to the lack of local data.  Local adaptations make better use of the
available data, improving accuracy of the model.

``Head'' participants (i.e., those with high BASIC-FED accuracy) also have
few, simple words, but their sentences are very different from the other
participants: e.g., ``gucci gang gucci gang gucci gang.''  Therefore,
(a) their local models outperform the unadapted federated model, and (b)
local adaptations improve accuracy for them, but not as much as for the
``tail'' participants.



\paragraphbe{Some participants never recover accuracy.}
In our image classification experiments, adapted models are always more
accurate than the local models regardless of the aggregation method.  In
the word prediction experiments, however, adapted models never reach the
same accuracy as the local models of some participants, especially with
ROBUST-FED.  We conjecture that median aggregation~\cite{yin2018byzantine}
prevents these participants from contributing to the federated model
at all.  As a consequence, the federated model is so bad for these
participants than when it is used to initialize local adaptation, the
final adapted model still has poor accuracy~\cite{grosse2019adversarial,
hanin2018start}.

\paragraphbe{Cumulative benefit of different adaptations.}
\autoref{fig:text_cai} shows cumulative improvement due to different
adaptations.  For BASIC-FED, the simplest FB technique performs best.
For DP-FED and ROBUST-FED, MTL performs better for the ``tail''
participants.


\section{Conclusion}
\label{sec:conclusion}

Federated learning is a promising approach to large-scale model training
on sensitive data.  Unfortunately, differential privacy and robust
aggregation reduce accuracy of federated models below that of the locally
trained models of many participants, removing their main incentive to
join federated learning.  We showed how local adaptation techniques
based on fine-tuning, multi-task learning, and knowledge distillation
help improve the accuracy of private and robust federated models for
individual participants, enabling them to reap the benefits of federated
learning.






\newpage
\bibliography{main}
\bibliographystyle{icml2020}
\balance
\newpage
\appendix
\section{Additional experiments}

\subsection{Adapting, then aggregating again}

To investigate whether it is beneficial to aggregate the adapted models
yet again, we use BASIC-FED on image classification.  We first train a
conventional federated model for 200 epochs with the learning rate of 0.1
and 2 internal epochs per participant, reaching 90.44$\%$ test accuracy.
We adapt by fine-tuning with the learning rate of 0.001 for 5, 50,
or 100 epochs.  Averaging the adapted models produces federated models
whose test accuracy is, respectively, 91.15$\%$, 92.64$\%$, and 89.22$\%$.

With the right learning rate and number of epochs, aggregating adapted
models can potentially produce a more accurate federated model\textemdash
at the cost of significantly increasing the training time for each
participant.  We leave an exploration of these tradeoffs to future work.

\subsection{Removing disincentivized participants}

As shown in \autoref{sec:fedworselocal}, there are 7,377 participants
in the word-prediction task whose local models are more accurate on
their data than the federated model and who thus have no incentive
to participate.  If we re-train the federated model on the remaining
72,623 participants, it achieves mean/median accuracy of 20.008$\%$ /
19.570$\%$ vs., respectively, 20.021$\%$ / 19.563$\%$ achieved by the
original model on 80,000 participants.  The new model performs well
even on the removed 7,377 participants, with mean accuracy of 20.076$\%$
vs.\ 20.301$\%$ for the original.  Among the 72,623 participants used to
train both models, the new model underperforms the original only on 974
(1.34$\%$) participants.

As discussed in \autoref{subsec:adaptionanalysis}, the removed
participants have (a) simpler and fewer words, and (b) their sentences are
outliers, very different from the rest of the participants.  We conjecture
that after removing these participants, the remaining set is more regular
yet sufficiently complex to train a model that performs comparably to
the original model.



\begin{figure}[!htb]
    \begin{tabular}{cc}
    \begin{minipage}[t]{0.8\linewidth}
        \includegraphics[width = 1\linewidth]{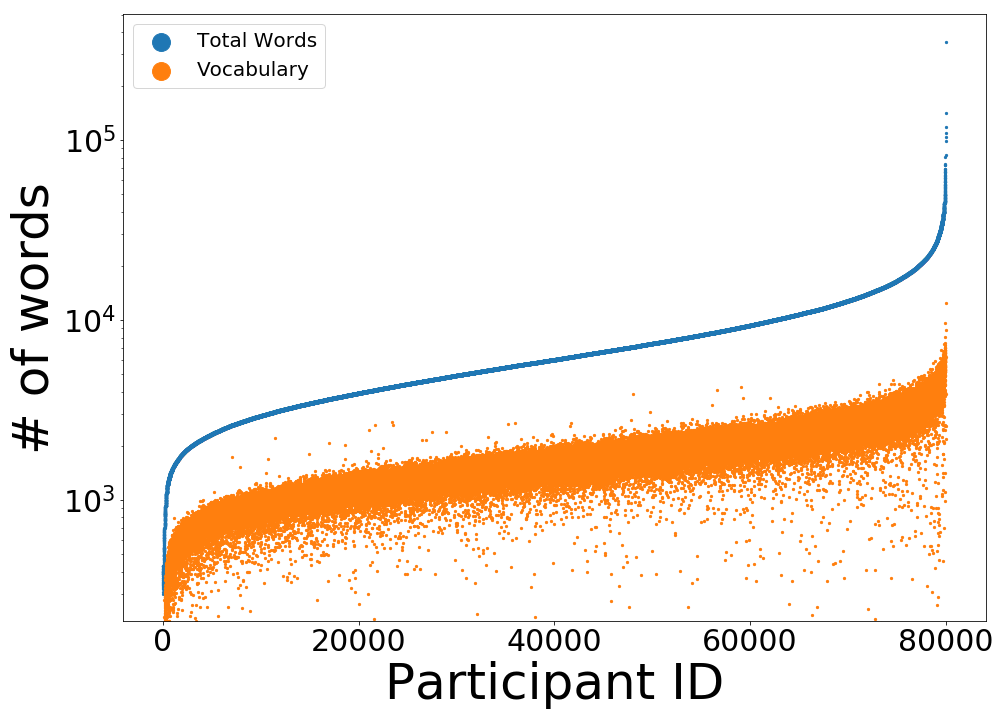}
    \end{minipage}
    \end{tabular}
    \begin{tabular}{cc}
    \begin{minipage}[t]{0.8\linewidth}
        \includegraphics[width = 1\linewidth]{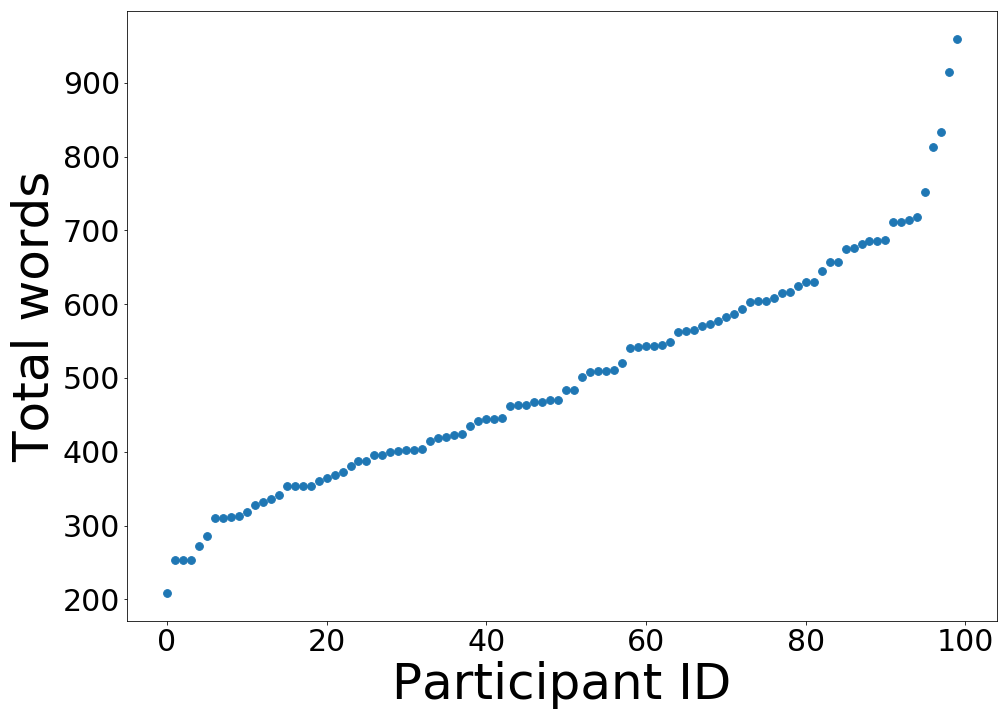}
    \end{minipage}
    \end{tabular}
    \caption{Size and complexity of participants' local data. Top: word prediction; Bottom: image classification.}
    \label{fig:data_imbalance}
    \end{figure}

\subsection{Imbalance of participants' local datasets}
\label{sec:imbalance}

We measure the imbalance between participants' local datasets.
\autoref{fig:data_imbalance} (Top) shows the size (total words) and
complexity (vocabulary size) of each participant's local data for the
word prediction task.  \autoref{fig:data_imbalance} (Bottom) shows the
size (total images) of each participant's local data for the image
classification task.

\end{document}